\documentclass[12pt]{article}
\usepackage{tgtermes, inputenc, amsmath, graphicx, hyperref, titling, url}
\usepackage[margin=1in]{geometry}
\usepackage{algorithm,algorithmic}
\usepackage{diagbox}
\usepackage{setspace}
\DeclareMathOperator*{\argmin}{argmin} 

\usepackage{natbib}

\title{%
    A study of tree-based methods and their combination \\
    \large Master's Project}
\author{Yinuo Zeng}
\date{June 2020}

\begin{document}

\maketitle

\begin{abstract}
Tree-based methods are popular machine learning techniques used in various fields. In this work, we review their foundations and a general framework the importance sampled learning ensemble (ISLE)  that accelerates their fitting process. Furthermore, we describe a model combination strategy called the adaptive regression by mixing (ARM), which is feasible for tree-based methods via ISLE. Moreover, three modified ISLEs are proposed, and their performance are evaluated on the real data sets. 
\end{abstract}

\section{Introduction}
With the increase of data volume and the continuous development in deep learning, although more and more traditional machine learning techniques are outperformed by artificial neural networks, tree-based methods are still popular. Random forest \citep{rf} is commonly used as a benchmark to evaluate the performance of nonparametric models, while XGBoost \citep{xgboost} performs well in Kaggle competitions and often competes with artificial neural networks.

Also, instead of relying on a specific method, people prefer to make decisions based on a combination of multiple models, which shows a better performance than a single one. Therefore, identifying the importance of each model by weights assignment is critical.

In this work, we delve into tree-based methods and study the importance sampled learning ensemble (ISLE) proposed by \cite{isle}, which accelerates the fitting process so that tree-based methods can be performed on large data sets within a reasonable time. In addition, under ISLE procedure, it is feasible to conduct the adaptive regression by mixing (ARM) \citep{arm} to assign weights among tree-based methods.

In Section 2, we review tree-based methods. Section 3 shows that these models can be viewed as an ISLE, and different variants of ISLE are proposed. Section 4 describes ARM. Real data experiments are presented in Section 5, and we end with a discussion in Section 6.

\section{Tree-based Methods}
\subsection{Decision Trees}
A decision tree \citep{tree} is the basic unit of tree-based methods, which divides the feature space into disjoint regions, and in each region, the mean of response is used as the predicted value for regression problems, while the majority vote of response is applied for classification problems. Specifically, for regression problems, $M$ disjoint regions $R_1, ..., R_M$ are partitioned from the feature space and a decision tree model has the form $f(x) = \sum_{m=1}^M c_m I(x \in R_m)$, where $c_m = avg(y_i | x_i \in R_m)$ is the average of response $y_i$ associated with predictor $x_i$ in region $R_m$.

A decision tree is a popular machine learning technique because its mechanism is easy to understand and can be interpreted. However, one of its significant disadvantages is instability. A small change in data would cause a considerable variation in the corresponding prediction so that its performance is not as outstanding as the other machine learning methods.

\subsection{Bagged Trees and Random Forests}
In order to overcome the drawbacks of a single decision tree, bagged trees \citep{bagging} are proposed. A bagged tree is an ensemble of decision trees generated on bootstrapped datasets. In the regression context, bootstrapping is applied to the original data to generate $L$ bootstrapped datasets. Then, each of them is fitted by a decision tree, and finally, the predicted response of a bagged tree is obtained by taking the average of results of $L$ decision trees.

Random forests \citep{rf} are a further improvement of bagged trees. Each decision tree fits on a modified bootstrapped dataset where features are randomly selected. The model has the form $f(x) = \frac{1}{L} \sum_{b=1}^L f_b(x)$, where $L$ is the number of bootstrapped datasets and $f_b(x)$ is a decision tree fitted on the $b$th bootstrapped dataset with subsampled features. Random forests decorrelate decision trees by alleviating the effect of strong features in the fitting process to increase prediction accuracy.

\subsection{MART}
Boosting is a powerful ensemble learning method that combines multiple base learners (an algorithm that gives a slightly better prediction accuracy than a random guess) to form a strong committee, which drastically improves performance. 
Adaboost \citep{adaboost} is the first practical boosting algorithm that was proposed to solve binary classification problems. Its popularity is from excellent predictability and impressive theoretical foundations. \cite{fsam} explained the mechanism of Adaboost from a statistical point of view via the forward stagewise additive modeling with exponential loss. Gradient boosting model (GBM), a general version of Adaboost, was proposed by \cite{gbm}, which enables the loss function to be any differentiable function.

Multiple Additive Regression Trees (MART) are GBM, whose base learners are decision trees. In order to further improve its prediction accuracy, shrinkage \citep{gbm} and subsampling \citep{sgbm} techniques are introduced. Specifically, in \textit{Algorithm 1}, subsampling is performed by random permutation of the training data (line 3) and selection of $\Tilde{N} < N$ observations (line 4). In line 11, $\nu \in (0, 1]$ is a shrinkage parameter.

\begin{algorithm}
  \caption{MART}
  \begin{algorithmic}[1]
    \STATE $F_0(x) = \argmin_\gamma \sum_{i=1}^N L(y_i, \gamma)$
  
    \FOR{$m$ = 1 to $M$}
      \STATE $\{\pi(i) \}_1^N$ = rand\_perm $\{ i \}_1^N$
      \FOR{$i$ = 1 to $\Tilde{N}$}
        \STATE $\Tilde{y}_{\pi(i) m} = - \left[ \frac{\partial L(y_{\pi(i)}, F(x_{\pi(i)}))}{\partial F(x_{\pi(i)})} \right]_{F = F_{m-1}}$
      \ENDFOR
      \STATE Fit a regression tree on samples $\{(\Tilde{y}_{\pi(i) m}, x_{\pi(i)})\}_{i=1}^{\Tilde{N}}$ and obtain regions $R_{j m}, j=1,...,J_m$
      \FOR{$j=1,...,J_m$}
        \STATE $\gamma_{j m} = \argmin_\gamma \sum_{x_{\pi(i) \in R_{j m}}} L(y_{\pi(i)}, F_{m-1}(x_{\pi(i)}) + \gamma)$
      \ENDFOR
      \STATE $F_m(x) = F_{m-1}(x) + \nu \sum_{j=1}^{J_m} \gamma_{j m} I(x \in R_{j m})$
    \ENDFOR
  \end{algorithmic}
\end{algorithm}

\section{ISLE}
\subsection{ISLE}
Although random forests and MART appear quite different, their tree-based generating procedure can be presented under a unifying framework. Inspired by numerical integration with importance sampling, a two-stage algorithm importance sampled learning ensemble (ISLE) \citep{isle} has been proposed.

In the first stage, candidate trees are generated.

\begin{algorithm}
  \caption{Ensemble Generation}
  \begin{algorithmic}[1]
    \STATE $F_0(x) = \argmin_\alpha \sum_{i=1}^N L(y_i, \alpha)$
  
    \FOR{$m$ = 1 to $M$}
      \STATE $\gamma_m = \argmin_\gamma \sum_{i \in S_m(\eta)} L(y_i, F_{m-1}(x_i) + f(x_i; \gamma))$
      \STATE $b_m(x) = f(x;\gamma_m)$
      \STATE $F_m(x) = F_{m-1}(x) + \nu b_m(x)$
    \ENDFOR
    \STATE ensemble = \{$b_1(x), b_2(x),... , b_M(x)$\}
  \end{algorithmic}
\end{algorithm}

In \textit{Algorithm 2}, $S_m(\eta)$ refers to a subsample of $N \times \eta$ ($\eta \in (0,1] $) of training observations that are randomly drawn without replacement (line 3), and in line 5, $\nu \in [0,1]$ is a shrinkage parameter.

Random forests and MART are special cases of this ensemble generation. Specifically, random forests can be viewed as this procedure by setting $\nu = 0$ and $\eta < \frac{1}{2}$ \citep{esl}, while MART follow the exact procedure by comparing \textit{Algorithm 2} with \textit{Algorithm 1}.

In the second stage, which is called post-processing, a linear model with lasso penalty \citep{lasso} is applied to reduce the number of trees without losing much prediction accuracy. 

$$\beta(\lambda) = \argmin_\beta \sum_{i=1}^N L(y_i, \beta_0 + \sum_{m=1}^M \beta_m b_m(x_i)) + \lambda \sum_{m=1}^M |\beta_m|$$

The penalty is reasonable since random forests and MART commonly generate a large number of trees $m = 1, ..., M$ so that modified dataset $b_m(x_i)$ is high-dimensional.

ISLE provides another perspective of tree-based methods. \cite{isle} demonstrated that in the ensemble generation stage, if trees are sufficiently diverse, then the post-processing improves the prediction accuracy remarkably. As shown in Section 5, the performance of ISLE is able to compete with fine-tuned random forests and MART with a much less cost of time.

\subsection{Modified ISLE}
Based on ISLE framework, we attempt to extend the post-processing stage and delve into different penalties. Although the lasso penalty is popular for feature selection, it does not handle highly correlated features well and does not provide a consistent estimate of coefficients. Therefore, we perform \textit{(1)} adaptive lasso \citep{alasso}, \textit{(2)} elastic net \citep{enet}, and \textit{(3)} adaptive elastic net \citep{aenet} penalties and hope they can overcome disadvantages of lasso penalty and increase prediction accuracy.

The post-processing becomes

\begin{equation}
\beta(\lambda, \gamma) = \argmin_\beta \sum_{i=1}^N L(y_i, \beta_0 + \sum_{m=1}^M \beta_m b_m(x_i)) + \lambda \sum_{m=1}^M |\hat{\beta}_{m}^{(lasso)}|^{-\gamma} |\beta_m| \ ,
\end{equation}

\begin{equation}
\beta(\lambda, \alpha) = \argmin_{\beta} \sum_{i=1}^N L(y_i, \beta_0 + \sum_{m=1}^M \beta_m b_m(x_i)) + \lambda \sum_{m=1}^M (\alpha \beta_m^2 + (1-\alpha)|\beta_m|) \ ,
\end{equation}

\begin{equation}
\beta(\lambda, \alpha, \gamma) = \argmin_\beta \sum_{i=1}^N L(y_i, \beta_0 + \sum_{m=1}^M \beta_m b_m(x_i)) + \lambda \sum_{m=1}^M (\alpha \beta_m^2 + (1-\alpha) |\hat{\beta}_{m}^{(enet)}|^{-\gamma} |\beta_m|) \ ,
\end{equation}

where $\lambda, \alpha$ and $\gamma$ are hyperparameters that are estimated by cross-validation. $\{ \hat{\beta}_{m}^{(lasso)} \}_{m=1}^M$ and $\{ \hat{\beta}_{m}^{(enet)} \}_{m=1}^M$ are coefficients of models with lasso and elastic net penalties, respectively.

\section{ARM}

The model combination is a strategy that combines multiple models by adaptively weighting them to make more precise and robust decisions. Instead of relying on a specific model, model combination commonly shows better performance and are used in reality frequently. However, one crucial problem of model combination is identifying the appropriate weight for each model. 

Adaptive regression by mixing (ARM) \citep{arm} is a feasible and flexible algorithm that can estimate weights in model combination for both parametric and nonparametric models. Also, it has been shown to be optimal in rates of convergence theoretically. However, this algorithm is computationally intensive, making the weights identification of nonparametric models very difficult.

Random forests and MART, as complex nonparametric models, have a time-consuming tuning procedure so that it is almost impossible to estimate their weights by ARM. However, ISLE provides another way to implement tree-based methods within a reasonable time so that the estimation of the weights in their model combination via ARM becomes feasible.

\begin{algorithm}
  \caption{ARM}
  \begin{algorithmic}[1]
    \STATE Initialize weights $w_j = 0, j=1,...,J$
    \FOR{m = 1 to M}
        \STATE Permute and split data into three parts $z^{(1)} = (x_i, y_i)_{i=1}^{n_1}$, $z^{(2)} = (x_i, y_i)_{i=n_1+1}^{n_1+n_2}$, and $z^{(3)} = (x_i, y_i)_{i=n_1+n_2+1}^n$. Let $n_3 = n-n_1-n_2$
    
        \STATE Obtain estimates $\hat{f}_{j, n_1}(x; z^{(1)})$ of $f$ based on $z^{(1)}$
    
        \STATE Estimate variance $\sigma^2$ by $\hat{\sigma}^2_j = \frac{1}{n_2} \sum_{i=n_1 + 1}^{n_1+n_2} (y_i - \hat{f}_{j,n_1}(x_i))^2$
    
        \STATE Compute $e_j = \frac{\prod_{i=n_1+n_2+1}^n h((y_i - \hat{f}_{j, n_1}( x_i))/\hat{\sigma}_{j})} {\hat{\sigma}_{j}^{n_3}}$, where $h$ is a density function
    
        \STATE Compute accumulated weights $w_{j} = w_{j} + \frac{e_{j}}{\sum_{l=1}^J e_l}$

    \ENDFOR
    \STATE The estimated weights are $\hat{w}_j = \frac{w_j}{M}$

    \STATE The final estimator is $\Tilde{f}_n(x) = \sum_{j=1}^J \hat{w}_{j} \hat{f}_{j,n}(x)$
  \end{algorithmic}
\end{algorithm}

\section{Results}
In this section, 17 models are compared and evaluated based on prediction accuracy in terms of mean squared error (MSE) and time costs. They are 
\textit{(1)} fine-tuned random forest (RF1), 
\textit{(2)} random forest from ensemble generation of ISLE (RF2), 
\textit{(3)} post-processing on RF2 with lasso penalty (RF\_lasso), 
\textit{(4)} post-processing on RF2 with adaptive lasso penalty (RF\_alasso), 
\textit{(5)} post-processing on RF2 with elastic net penalty (RF\_enet), 
\textit{(6)} post-processing on RF2 with adaptive elastic net penalty (RF\_aenet), 
\textit{(7)} fine-tuned MART (MART1), 
\textit{(8)} MART from ensemble generation of ISLE (MART2), 
\textit{(9)} post-processing on MART2 with lasso penalty (MART\_lasso), 
\textit{(10)} post-processing on MART2 with adaptive lasso penalty (MART\_alasso), 
\textit{(11)} post-processing on MART2 with elastic net penalty (MART\_enet)), 
\textit{(12)} post-processing on MART2 with adaptive elastic net penalty (MART\_aenet), 
\textit{(13)} model combination of RF1 and MART1 by ARM (ARM\_tree), 
\textit{(14)} model combination of RF\_lasso and MART\_lasso by ARM (ARM\_lasso), 
\textit{(15)} model combination of RF\_alasso and MART\_alasso by ARM (ARM\_alasso),
\textit{(16)} model combination of RF\_enet and MART\_enet by ARM (ARM\_enet), and 
\textit{(17)} model combination of RF\_aenet and MART\_aenet by ARM (ARM\_aenet).

Considering the computation time, we simplify the tuning process of models and select the best model by 5-fold CV. For RF1, the number of randomly selected features $m$ is the only hyperparameter, while RF2 has $m = p/3$ \citep{esl}, where $p$ is the total number of predictors. In addition, the tuning parameters of MART1 are the depth of a tree, the number of trees, and the shrinkage coefficient. For MART2, we refer to suggestions of \cite{isle} and set tree depth = 2, subsampling ratio = 0.2, and shrinkage coefficient = 0.01 to enable trees to be sufficiently different. The number of trees in RF2 and MART2 is chosen according to the size of the data. Moreover, a reasonable amount of hyperparameters are selected for post-processing without loss of generality. Furthermore, in model combination via ARM, training data is divided into three parts in the ratio of 2:1:1 (line 3 in \textit{Algorithm 3}), $h$ is chosen to be the Gaussian density (line 6) \citep{arm}, and $M=20$ iterations are performed (line 2) considering its intensive computation.

These models are evaluated on four datasets: \href{https://cran.r-project.org/web/packages/MASS/MASS.pdf}{Boston} data frame obtained from R package MASS includes 506 observations with 13 features, while there are 442 samples and 10 variables in \href{https://cran.r-project.org/web/packages/lars/lars.pdf}{Diabetes} dataset that is from R package lars. The other two datasets containing more than 1000 samples are from the UCI repository, \href{https://archive.ics.uci.edu/ml/datasets/Appliances+energy+prediction}{Energy} data has 2000 examples and 27 features, while \href{https://archive.ics.uci.edu/ml/datasets/Beijing+PM2.5+Data}{AirQuality} dataset includes 3000 observations with 9 predictors.

\begin{table}
\caption{Prediction Accuracy (MSE) and Computation Time (in seconds) of Models}
\centering
\begin{tabular}{|l|l|l|l|l|}
\hline
\backslashbox{Model}{Dataset}     & Boston          & Diabetes           & Energy              & AirQuality          \\ \hline
RF1                     & 10.14 (13) & 3079 (9)   & 4194 (268)  & 4668 (95)   \\
RF2             & 29.02      & 3615       & 10264       & 7330        \\
RF\_lasso    & 8.89 (1)   & 3123 (2)   & 4812 (11)   & 5069 (23)   \\
RF\_alasso    & 9.06 (6)   & 3158 (7)   & 4885 (40)   & 5190 (80)   \\
RF\_enet     & 8.77 (6)   & 3095 (7)   & 4623 (39)   & 5053 (81)   \\
RF\_aenet     & 9.10 (11)  & 3176 (12)  & 4851 (69)   & 5176 (138)  \\
\hline
MART1                   & 6.97 (40)  & 3037 (41)  & 3502 (206)  & 4582 (164)  \\
MART2            & 9.00       & 4382       & 6604        & 5674        \\
MART\_lasso   & 7.94 (5)   & 3214 (5)   & 4390 (24)   & 4988 (28)   \\
MART\_alasso  & 9.08 (13)  & 3241 (12)  & 4583 (114)  & 5040 (112)  \\
MART\_enet    & 7.90 (12)  & 3140 (12)  & 4378 (57)   & 4993 (85)   \\
MART\_aenet   & 11.39 (21) & 3896 (21)  & 5807 (135)  & 5017 (152)  \\
\hline
ARM\_tree       & 7.37 (546) & 3017 (534) & 3549 (4294) & 4532 (2428) \\
ARM\_lasso     & 8.30 (137) & 3112 (114) & 4138 (403)  & 4911 (586)  \\
ARM\_alasso    & 8.53 (285) & 3102 (258) & 4238 (1398) & 4943 (1892) \\
ARM\_enet      & 8.22 (282) & 3056 (259) & 4050 (893)  & 4896 (1602) \\
ARM\_aenet     & 8.95 (445) & 3262 (399) & 4773 (1731) & 4931 (2688) \\ \hline
\end{tabular}
\end{table}

\textit{Table 1} shows MSE and time costs (in the parentheses) of 17 models on the four datasets. Since randomness is introduced in the fitting process of random forests and MART, we repeat the whole procedure 10 times and MSE and time costs in \textit{Table 1} are the average.

In addition to RF1, MART1, and  ARM\_tree, which have the smallest MSE respectively, it is evident that RF\_enet, MART\_enet, and ARM\_enet have the closest MSE to them, while RF\_lasso, MART\_lasso, and ARM\_lasso have the third smallest MSE. 

When it comes to computation time,  post-processing with lasso penalty (RF\_lasso, MART\_lasso, and ARM\_lasso) is the most computationally efficient algorithm, and elastic net penalty (RF\_enet, MART\_enet, and ARM\_enet) is the second although the time costs of adaptive lasso penalty (RF\_alasso, MART\_alasso, and ARM\_alasso) is competitive with it on specific datasets. Additionally, the advantage of ISLE is shown in fitting ARM. In general, the computation time of ARM combing ISLEs (ARM\_lasso, ARM\_alasso, ARM\_enet, and ARM\_aenet) is much less than ARM\_tree that is even impossible to implement when datasets become large.

Also, \textit{Table 1} demonstrates the properties of ISLE. In the ensemble generation stage (RF2 and MART2), although the prediction accuracy is undesirable compared with RF1 and MART1, trees can be generated very fast. The post-processing stage significantly improves the performance (for example, RF\_lasso and MART\_lasso) and can compete with RF1 and MART1.

\section{Discussion}
In this project, we focus on two popular tree-based methods, random forests and MART, and attempt to accelerate their fitting process via a general framework ISLE. A comprehensive model comparison is performed, and their prediction accuracy and computation time are evaluated. Lasso and elastic net penalties are suggested in the post-processing stage of ISLE, which can reduce fitting time remarkably without losing much performance. Specifically, if time-saving is the priority, then the lasso penalty is preferred, while the elastic net penalty is recommended when prediction accuracy is considered. However, it seems that adaptive penalties are not desirable choices for ISLE. Moreover, the model combination of random forest and MART via ARM can be implemented via ISLE within a reasonable time.

However, this work is limited. We only study ISLE in the regression context and do not take classification problems into account. Also, it is worthwhile to delve into other penalties for post-processing as well as different loss functions for evaluation. Additionally, XGBoost, a powerful tree-based method, should be studied from an ISLE perspective to decrease the computation time. These problems will be investigated in future research.















\newpage

\bibliography{ref}
\bibliographystyle{apalike}

\end{document}